\documentclass[10pt,twocolumn,letterpaper]{article}

\usepackage{cvpr}              

\usepackage{graphicx}
\usepackage{amsmath}
\usepackage{amssymb}
\usepackage{booktabs}
\usepackage{amssymb}
\usepackage[accsupp]{axessibility} 

\usepackage[pagebackref,breaklinks,colorlinks]{hyperref}

\usepackage[capitalize]{cleveref}
\crefname{section}{Sec.}{Secs.}
\Crefname{section}{Section}{Sections}
\Crefname{table}{Table}{Tables}
\crefname{table}{Tab.}{Tabs.}

\def\cvprPaperID{1} 
\def\confName{CVPR}
\def\confYear{2023}

\begin{document}

\title{Shape-Net: Room Layout Estimation from Panoramic Images Robust to Occlusion using Knowledge Distillation with 3D Shapes as Additional Inputs}

\author{
Mizuki Tabata$^1$
\and
Kana Kurata$^2$
\and
Junichiro Tamamatsu$^1$
\and
$^1$\text{NTT Access Network Service Systems Laboratories},
$^2$\text{NTT Human Informatics Laboratories} \\
{\tt\small \{mizuki.tabata, kana.kurata, junichirou.tamamatsu\}@ntt.com}
}

\maketitle

\begin{abstract}
   Estimating the layout of a room from a single-shot panoramic image is important in virtual/augmented reality and furniture layout simulation. This involves identifying three-dimensional (3D) geometry, such as the location of corners and boundaries, and performing 3D reconstruction. However, occlusion is a common issue that can negatively impact room layout estimation, and this has not been thoroughly studied to date. It is possible to obtain 3D shape information of rooms as drawings of buildings and coordinates of corners from image datasets, thus we propose providing both 2D panoramic and 3D information to a model to effectively deal with occlusion. However, simply feeding 3D information to a model is not sufficient to utilize the shape information for an occluded area. Therefore, we improve the model by introducing 3D Intersection over Union (IoU) loss to effectively use 3D information. In some cases, drawings are not available or the construction deviates from a drawing. Considering such practical cases, we propose a method for distilling knowledge from a model trained with both images and 3D information to a model that takes only images as input. The proposed model, which is called Shape-Net, achieves state-of-the-art (SOTA) performance on benchmark datasets. We also confirmed its effectiveness in dealing with occlusion through significantly improved accuracy on images with occlusion compared with existing models.
\end{abstract}

\begin{figure}[t]
\centering
\includegraphics[width=1.0\linewidth]{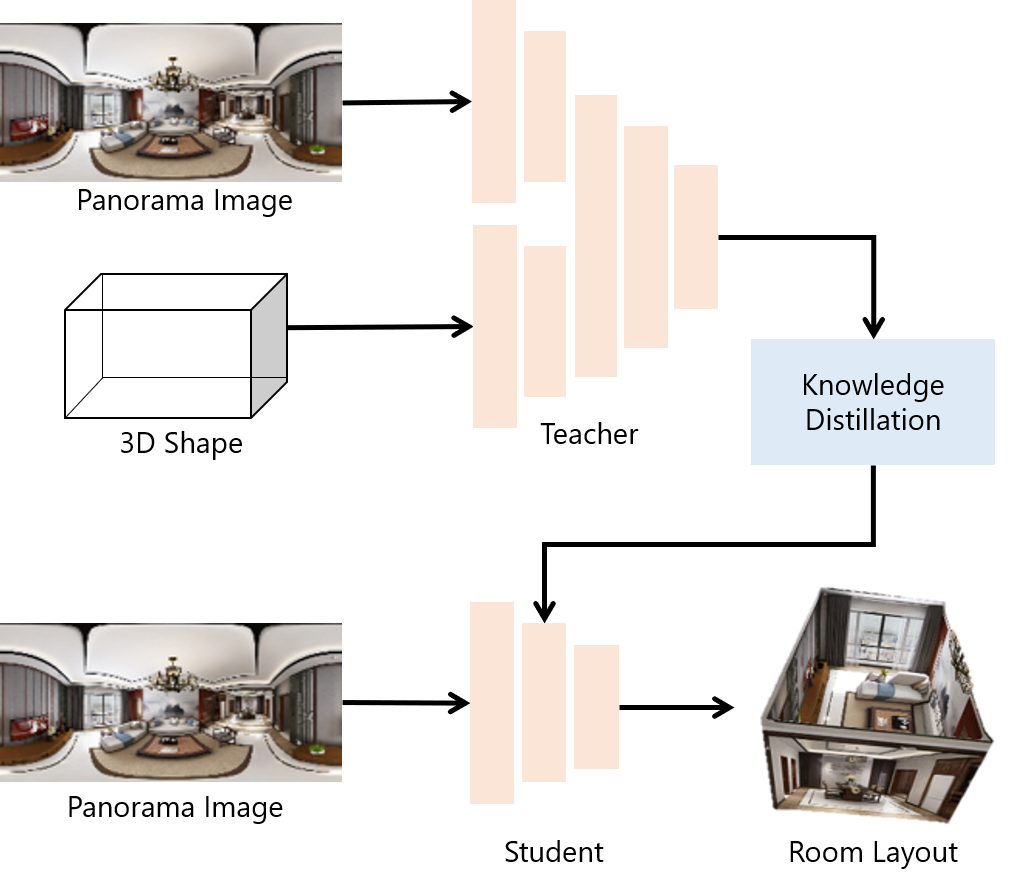}
\caption{Overall architecture of Shape-Net. Teacher model is trained in advance by providing 3D shape with images, and extracted features of training model are used while student model is trained with only images as input. Trained student model performs inference of room layouts.}
\label{fig:teaser}
\end{figure}
\section{Introduction}
\label{sec:intro}

Room layout estimation from panoramic images is widely used for 3D room modeling, including applications in virtual reality, augmented reality, and furniture arrangement. The method estimates the positional relationships of the components of a room, e.g., corners, boundaries, and wall surfaces, even without directly estimating the layout, on the basis of the Manhattan World assumption~\cite{Manhattan_world_assumption} that all walls are orthogonal. Since the geometric information of room components varies among different room types, deep learning has been successful in solving this problem. In recent years, the development of deep neural networks has facilitated remarkable advancements in the estimation of room layouts from a single panorama image~\cite{LayoutNet, DulaNet, HorizonNet, HoHoNet, Led2Net, LGTNet}. From the perspective of data capacity, the method of estimating layouts using a single panoramic image is a favored approach.

Occlusion frequently occurs in panoramic images of rooms since it is difficult to position a camera in such a way that all walls are visible in rooms with intricate shapes. Occlusion degrades the quality of layout estimation as it hides the room components, so it is one of the main issues to be addressed. Conventional approaches that solely rely on a 2D panoramic image face challenges to solve the occlusion issue, as there is a lack of compensating information behind the occlusion. Therefore, additional information is required to complement occluded areas. 3D shape information, such as drawings contained in layout data, can be used to supplement data surrounding an occlusion. Using 3D information for occluded regions raises two issues: 1) providing 3D shape information to a model does not guarantee its effective use, particularly for areas where occlusion occurs, and 2) although the dataset used for training includes 3D information, in practice, there may be cases where the drawings are unavailable, or reconstruction has resulted in deviations from the drawings. To address these issues, we propose Shape-Net: a Transformer\cite{ViT}-based knowledge distillation model with a novel 3D IoU loss function (\cref{fig:teaser}).

To solve the first issue, 3D IoU loss is introduced into the proposed model. The loss calculates the IoU of the ground truth and estimated room shape to take into account the volume of the room and does not decrease that much for areas of occlusion, which works well on occlusion. The knowledge distillation model resolves the second issue, as a student model enables inference from the input of only images while incorporating the training results from a teacher model that takes both images and 3D shapes as inputs. Shape-Net outperforms state-of-the-art (SOTA) models on benchmark datasets. We also tested our model on one of the dataset consisting only of scenes with occlusion. The test results show that the model achieves the highest accuracy, and the difference in accuracy increases compared with that on the dataset without occlusion. This indicates that Shape-Net is robust to occlusion.

The network architecture of Shape-Net allows use in situations where a pair of an image and a drawing of a room does not match, or an image of a room differs from a drawing. In addition, the student model does not require layers to process 3D input, and it gains 3D information by knowledge distillation from the teacher model, which shortens the inference time.

The main contributions of this paper are summarized as follows:
\begin{itemize}
   \item We introduce a 3D IoU loss function for room layout estimation.
   \item We propose a knowledge distillation model that infers from a single-shot image while using the training results from the input of both images and 3D shape.
   \item We evaluate the proposed model on benchmark datasets and demonstrate its effectiveness in dealing with occlusion.
\end{itemize}

\section{Related Work}
We review studies on layout estimation from a single-shot panoramic image, knowledge distillation, and cross-modality since our study involves the cross-modality of a room image and its corresponding 3D data using knowledge distillation for room layout estimation.

\subsection{Layout Estimation}
Most studies have estimated room layouts by detecting room geometry, e.g., boundary probability maps of walls and corners~\cite{edgemap, LayoutNet, HorizonNet}, and wall-surface classes~\cite{zhao2017physics, im2cad, DulaNet}. Methods that detect wall boundaries and corners show higher accuracy than those that detect wall surfaces, and they have been used more frequently in recent years. While most of these studies use the loss of 2D pixel coordinates, LED$^{2}$-Net and LGT-Net consider 3D geometric information through differentiable depth rendering~\cite{Led2Net} and through depth/height loss~\cite{LGTNet}. Although these studies consider room geometry in 3D spaces, the losses around occlusion are still small as they account for only the horizontal length or vertical height independently. As a result, their models have difficulty compensating for largely occluded areas. IoU loss can overcome this problem because it calculates the volume of a room. Although IoU loss calculation for 3D bounding boxes was proposed for 3D object detection ~\cite{iouloss}, to our knowledge, no study has proposed 3D IoU loss for complex shapes such as L-shapes. We devised an IoU loss for complex shapes for estimating room layouts in this work.

\subsection{Knowledge Distillation}
In deep learning, models with a large number of layers and parameters typically show superior performance, albeit at an increased computational cost. To mitigate this issue, the method of knowledge distillation is used. In knowledge distillation, a high-accuracy model referred to as a teacher model is trained, and the knowledge gained is utilized to train a lightweight and easily deployable student model. This approach aims to produce models that are lightweight yet comparable in accuracy to their teacher models~\cite{FitNets, BornAgain, distilling, knowledge_concentration}.

Essentially, knowledge distillation uses the teacher model's output to train the student model. The method uses the teacher's output as a soft target for learning, such that the distribution of the student's output is similar to that of the teacher's output, while the training data labels are used as a hard target. Various methods of distilling knowledge have been proposed, including a method of ensuring that the output distribution of the student model is similar to that of the teacher model, as described above~\cite{BornAgain, distilling, knowledge_concentration}, and a method of using features from the middle layer as well as the teacher's output~\cite{FitNets, KD_middle}. Notably, the latter method is more effective in training deeper networks~\cite{kd_survey}. Some approaches use privileged information, such as descriptive text or human posture, which is fed to the teacher model during training, and the trained weights are utilized for the student model's training~\cite{KD_privileged}. In this study, we used the 3D shape as privileged information in the teacher model, and the student model was trained using the soft target loss of the features from the middle layer, as the proposed model has several feature processing modules.

\subsection{Cross-Modality}
 The fusion of information from different modalities has been actively studied in visual question answering~\cite{Dualnet, MCBilinear, MFBilinear, Co-attention}, and its applications are increasing beyond image and natural language processing. Various fusion methods have been explored, including concatenation~\cite{concat}, bilinear pooling~\cite{MCBilinear, MFBilinear}, and co-attention~\cite{MFBilinear, Co-attention}.

The fusion of 2D and 3D information has been extensively studied in 3D object detection, which involves the use of both images and point clouds as input~\cite{PointFusion, Joint3D, deep_continuous_fusion}. Moreover, the fusion of RGB (red, green, blue) images with depth images obtained through LiDAR (light detection and ranging) sensors has been examined~\cite{Lidar_Stereo_Fusion, deepfusion}. Given a rough alignment of these RGB and depth images, they are generally concatenated to create fused features~\cite{deepfusion}. Additionally, integrating shape features with image features through simple concatenation has been demonstrated to improve the accuracy of 3D object pose estimation, even in cases where the features are not aligned.~\cite{pose_from_shape}.

There have been some studies on cross-modality in indoor spaces, such as 2D-2D cross-modality for matching floor plans used by real estate companies with images of individual rooms~\cite{multi-modal_room} or 2D-3D cross-modality for mapping the texture images of manholes onto 3D models ~\cite{FPNet}. However, no studies have investigated the use of 2D-3D modalities for room layout estimation. Our study addresses this gap by exploring the potential of using these modalities for estimating room layouts.

\begin{figure*}
\begin{center}
\includegraphics[width=1.0\linewidth]{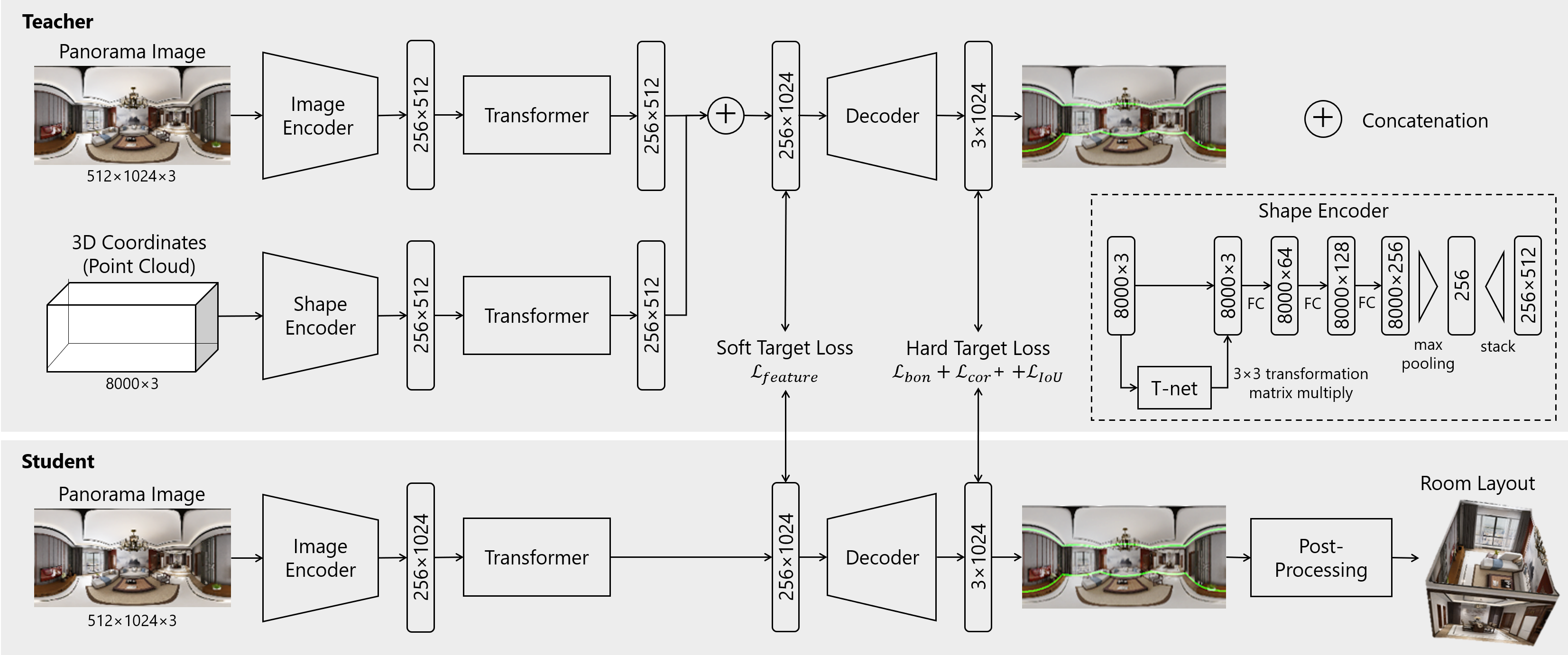}
\end{center}
   \caption{
   Network architecture of Shape-Net.
   To train student model, we use both the soft target and hard target loss. Former is computed on the basis of features extracted from middle layer of both teacher and student models, while the latter is determined from labels of datasets. After completion of training, student model is capable of inferring room layouts and reconstructing 3D layouts. Architecture of shape encoder in teacher model is also depicted.
   }
\label{fig:network_architecture}
\end{figure*}

\section{Methodology}
The proposed modes is a Transformer-based knowledge distillation model using 3D IoU loss. This section describes its network architecture and loss functions.
\subsection{Network Architecture}
The network architecture of our model is illustrated in \cref{fig:network_architecture}. It consists of a teacher and student model. First, the teacher model is trained with panoramic images and point clouds as inputs. Subsequently, the student model is trained with only panoramic images using the trained results from the teacher model to perform inference. These point clouds can be obtained by interpolating between points generated from ground truth image coordinates of corners. The methods of inputting point clouds are discussed in \ref {Input}.

The teacher model begins with extracting image features using the image encoder proposed in~\cite{HorizonNet} and shape features using a shape encoder based on Point-Net\cite{PointNet}. The input sizes are $512\times1024\times3$ (height, width, channel) for a panorama image and $n\times3$ for a point cloud. $n$ is the number of points, and was set to 8000 in this work. Both encoders produce a feature sequence $\mathbb{R}^{{N}\times{D}}$, where $N$ is 256, and $D$ is 512 in our implementation. In the shape encoder, the input points are multiplied by a $3\times3$ affine transformation matrix, which is regressed from a set of input points. They are fed into a multilayer perceptron with output sizes of 64, 128, and 256 and then transformed to $256\times1$ features using max pooling. To match the size of the image features obtained by the image encoder, the shape features are stacked 512 times in the dimensional direction. The image and shape features extracted from each encoder are processed by Transformer encoder layers~\cite{ViT} since global attention of Transformer~\cite{vit_vs_cnn} can capture the global relationships between spatially distant corners and wall boundaries, which leads to solve the occlusion problem. The Transformer encoder contains six multi-head self-attention layers~\cite{Attention} with eight heads each. The features are concatenated to the output $\mathbb{R}^{{N}\times{2D}}$. The feature sequence is processed by a decoder composed of bidirectional long-short-term-memory layers~\cite{LSTM} as in ~\cite{HorizonNet}. The student model lacks the shape encoder, and the size of image features is $\mathbb{R}^{{N}\times{2D}}$, otherwise the same network architecture as the teacher model. We used the post-processing method proposed in HorizonNet~\cite{HorizonNet}.

\begin{figure}[t]
  \centering
   \includegraphics[width=0.9\linewidth]{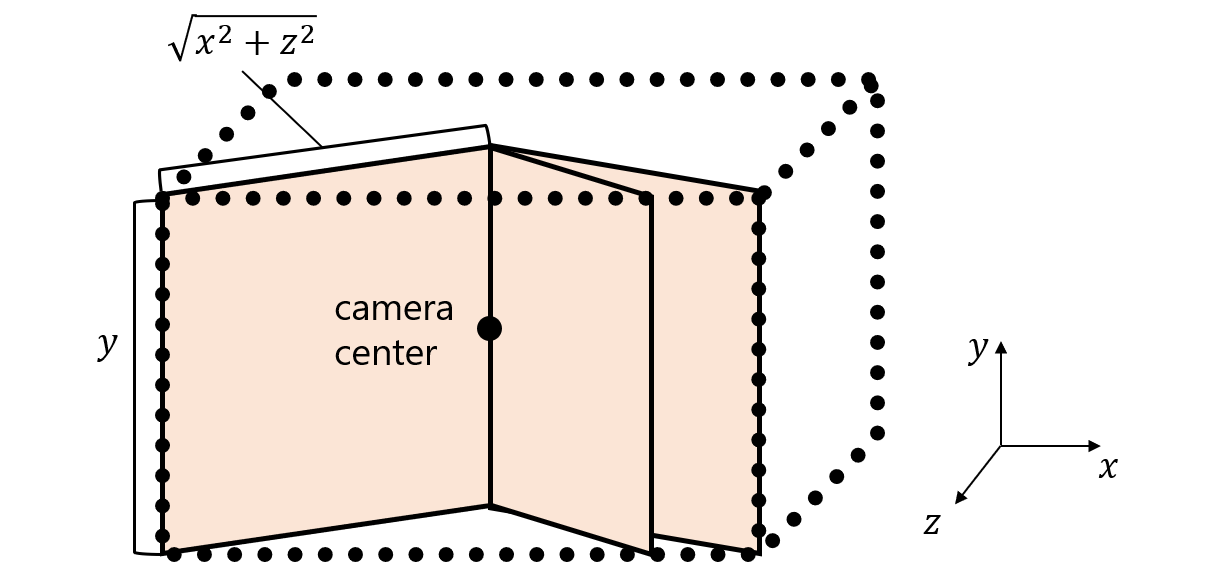}
   \caption{Illustration of IoU calculation for room layout estimation. We calculate IoU by integrating cross sections including camera center colored in pink. 3D coordinates in calculation follow coordinate system in illustration.}
   \label{fig:loss_function}
\end{figure}
\subsection{Loss Function}
We used the mean absolute error loss (L1 loss) for the image coordinates of the ceiling-wall and wall-floor boundaries, denoted as $\mathcal{L}_{b}$. To design a system to use a point cloud when occlusion occurs, the weight $\lambda$ is applied to $\mathcal{L}_{b}$ only in cases of occlusion. In this study, $\lambda$ was set to five when occlusion was present and one otherwise. The pixels for occlusion were defined to be vertical coordinates in the image of adjacent pixels that are more than five pixels apart in annotation. For the image coordinates of the corners, we used binary cross entropy loss $\mathcal{L}_{c}$.

Furthermore, we incorporated 3D IoU loss $\mathcal{L}_{IoU}$ to enable the model to account for the structure of a room in three dimensions for enhancing robustness against occlusion. We devised a calculation method that use the summation of cut-out rectangles from a room (\cref{fig:loss_function}). It can be used even for non-cuboid rooms. Initially, we projected the image coordinates of estimated wall boundaries into 3D space. The number of pixels $p_{w}$ is that of the 3D boundary coordinates between the ceiling and wall and between the wall and floor. We first calculated the cross-sectional area colored in pink in \cref{fig:loss_function} from the 3D coordinates $(x, y, z)$ generated by the ground truth and $(\hat{x}, \hat{y}, \hat{z})$ from the estimated boundaries. The cross-sectional area for each pixel is $V$ and $\hat{V}$, respectively.
\begin{equation}
  V = \sum_{p_{w}}y\sqrt{(x^{2}+z^{2})}
  \label{eq:volume}
\end{equation}
    The volume of the intersection is obtained by the following equation where $h \in \{y, \hat{y}\}$ and $w \in \{\sqrt{(x^{2}+z^{2})}, \sqrt{(\hat{x}^{2}+\hat{z}^{2})}\}$.
\begin{equation}
  V_{int} = \sum_{p_{w}}\min{hw}
  \label{eq:v_int}
\end{equation}
The volume of the union $V_{uni}$ is $V+\hat{V} - V_{int}$, and IoU is $V_{int}/V_{uni}$. $\mathcal{L}_{IoU}$ is expressed as follows.
\begin{equation}
  \mathcal{L}_{IoU} = 1-IoU
  \label{eq:iou_loss}
\end{equation}

The knowledge distillation uses the teacher's output as a soft target, while it uses the data labels as a hard target. The loss functions: $\mathcal{L}_{b}$, $\mathcal{L}_{c}$, and $\mathcal{L}_{IoU}$ are the hard target loss used in both the teacher and the student model. In addition to those loss functions, the student model uses the soft target loss $\mathcal{L}_{soft}$, which measures the L1 loss between the features after concatenation in the teacher model and those after the Transformer layers in the student model (\cref{fig:network_architecture}). The total loss function for the teacher model $\mathcal{L}_{T}$ and that for the student model $\mathcal{L}_{S}$ are calculated as:
\begin{equation}
  \mathcal{L}_{T} = \lambda\mathcal{L}_{b} + \mathcal{L}_{c}+\mathcal{L}_{IoU}.
  \label{equ:L_T}
\end{equation}
\begin{equation}
  \mathcal{L}_{S} = \lambda\mathcal{L}_{b} + \mathcal{L}_{c}+\mathcal{L}_{IoU} + \mathcal{L}_{soft}.
  \label{equ:L_S}
\end{equation}

\begin{figure*}
\begin{center}
\includegraphics[width=1.0\linewidth]{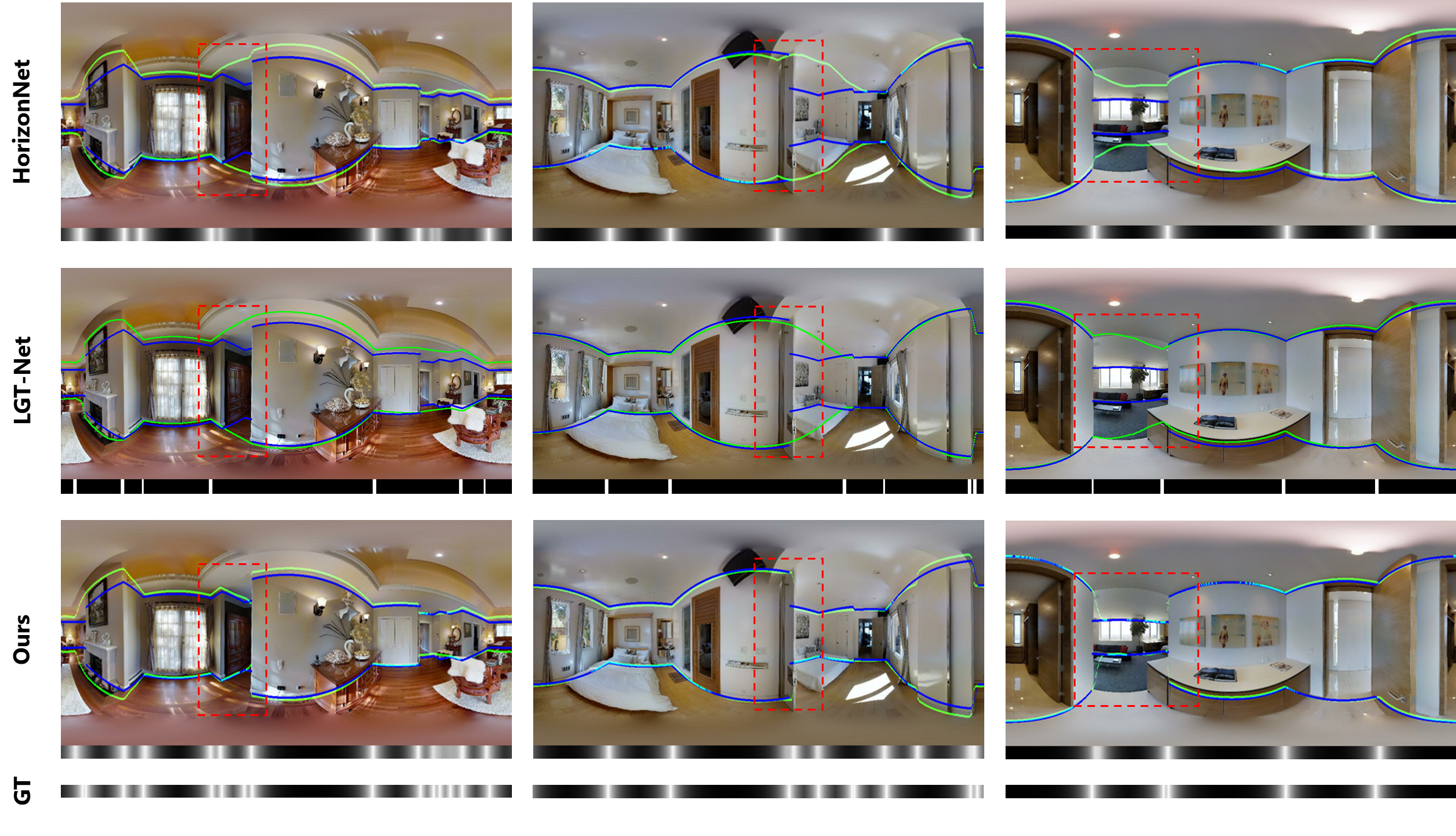}
\end{center}
   \caption{
   Qualitative results of layout estimation of general rooms without post-processing using HorizonNet~\cite{HorizonNet}, LGT-Net~\cite{LGTNet}, and our model on Matterport3D~\cite{Mt3D}. Blue lines in images show ground truth, while green lines show estimated results of wall boundaries. We also show estimated corner position in white at bottom of image and ground truth in bottom row. Corner locations estimated with LGT-Net were set at point that normal changes for convenience. Areas inside red dashed rectangles indicate where occlusion occurs, and proposed model succeeded in estimating boundaries.
   }
\label{fig:qualitative_results}
\end{figure*}

\section{Experiments}
\subsection{Implementation Details}
Shape-Net was implemented using PyTorch~\cite{pytorch}, and training was carried out with the Adam optimizer~\cite{Adam} using a batch size of four, learning rate of 0.0001, 1000 epochs on the Pano\_S2D3D~\cite{PanoContext, S2D3D} and Matterport3D~\cite{Mt3D} datasets, and 50 epochs on the Structured3D dataset~\cite{St3D}. Note that all the results discussed below are calculated by the student model of the best epoch in the validation split. We trained our model on an Nvidia GTX 2080 Ti and an Intel i79700 3.00-GHz CPU.

To augment the input data, we applied horizontal inversion, horizontal rotation, luminescence change, and Pano Stretch, which extends the images and annotations in the depth direction~\cite{HorizonNet}. For the input of the point cloud, we applied horizontal inversion and Pano Stretch. To enhance the model's robustness to occlusion, we also trained all models with occlusion-added images by applying the Cutout augmentation~\cite{Cutout}. Cutout masks an image with a black square region with fixed side lengths. We chose this augmentation method because it requires fewer model weights than others. To guarantee that critical areas for estimating room layouts were shaved off, we applied masks to images with side lengths of 50 pixels at three arbitrary corner positions in the images. The probability of applying Cutout was set to 50\%.

\subsection{Datasets}
We evaluated the proposed model on three datasets: a dataset that was mix of the PanoContext~\cite{PanoContext} and Stanford2D-3D~\cite{S2D3D} datasets (hereafter referred to as Pano\_S2D3D), Matterport3D~\cite{Mt3D}, and Structured3D~\cite{St3D}. Evaluation on the Pano\_S2D3D datasets has been performed by other models~\cite{HorizonNet, Led2Net, LGTNet}, thus we followed the composition of the dataset. While the Pano\_S2D3D and Matterport3D datasets~\cite{PanoContext, S2D3D, Mt3D} consist of real room data, Structured3D~\cite{St3D} is a synthetic dataset.

PanoContext~\cite{PanoContext} contains 514 room images and Stanford2D-3D~\cite{S2D3D} contains 552 room images. Pano\_S2D3D has only cuboid room layouts. We followed the data split offered by LayoutNet~\cite{LayoutNet}. The data split is composed of 817 pieces of training data, 79 pieces of validation data, and 166 pieces of test data.

Matterport3D~\cite{Mt3D} contains 2295 room layouts, including non-cuboid rooms. We followed the data split and annotation of LED$^{2}$-Net~\cite{Led2Net}. The data split consists of 1647 pieces of training data, 190 pieces of validation data, and 458 pieces of test data.

We evaluated methods for inputting point clouds using the Structured3D dataset~\cite{St3D}. Structured3D~\cite{St3D} contains 21835 room layouts, and more than 196k photo-realistic 2D renderings of the rooms. We followed the data split and annotation of HorizonNet~\cite{HorizonNet}. The data split consists of 18362 pieces of training data, 1776 pieces of validation data, and 1693 pieces of test data.

\subsection{Overall Performance}
We first evaluated our model for cuboid layouts on the Pano\_S2D3D dataset~\cite{PanoContext, S2D3D}, as presented in \cref{table:cuboid_results}, using the same evaluation metrics proposed in HorizonNet ~\cite{HorizonNet}: the IoU of 3D room layouts (3D IoU), corner error (CE), and pixel error (PE). The value of highest accuracy for each metric is shown in bold in all tables. Our model outperformed LGT-Net~\cite{LGTNet} by 1.06\% in 3D IoU and 3.08\% in CE. It also achieved almost the same accuracy in PE as HorizonNet~\cite{HorizonNet}. While PE and CE measure the average error between image coordinates of corners and boundaries, the 3D IoU is a metric for evaluating corner errors in 3D space; thus, a higher value in one measure does not necessarily entail a higher value in the other, leading to the above results.

\Cref{table:overall_performance} shows the performances of our model and those of conventional models for room layouts including non-cuboid rooms on Matterport3D~\cite{Mt3D}. As CE and PE are evaluation metrics for cuboid rooms, we evaluated the models with the metrics described in~\cite{review_room_layout}, i.e., 2D and 3D IoU for non-cuboid rooms. The results in \cref{table:overall_performance} indicate that our model outperformed all other models. Specifically, our model improved 2D IoU by 0.29\% and 3D IoU by 0.50\% compared with LGT-Net~\cite{LGTNet}. The qualitative results in \cref{fig:qualitative_results} indicate that our model made improvements in predicting occluded areas with fewer errors compared with other models.

\begin{figure}[t]
  \centering
   \includegraphics[width=1.0\linewidth]{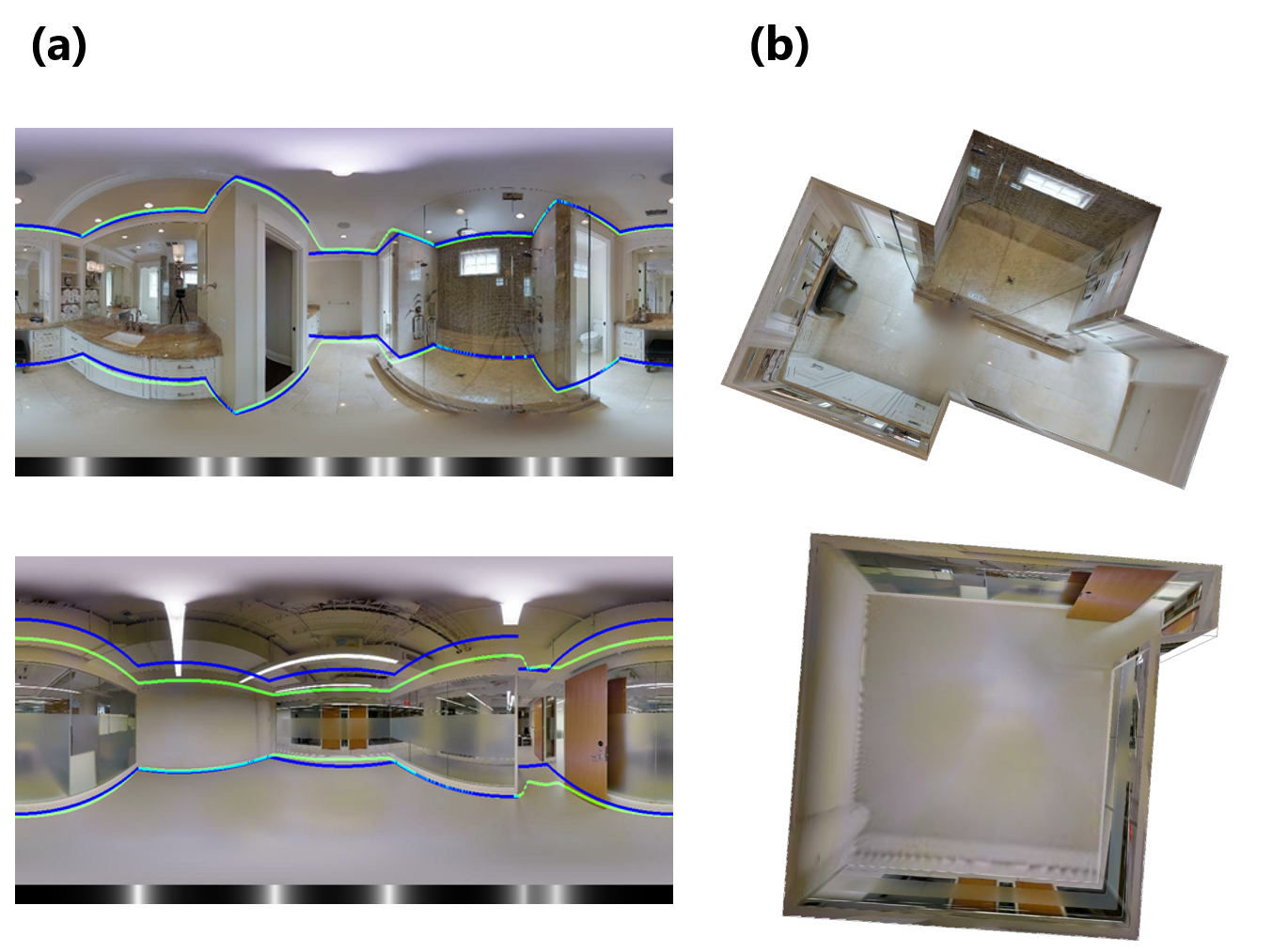}
   \caption{(a) Qualitative results of room layout estimation without post-processing. (b) Visualization of 3D layout corresponding to (a) images.}
   \label{fig:3D_layout}
\end{figure}

3D layouts reconstructed by our model are illustrated in \cref{fig:3D_layout}, exemplifying the precise reconstruction of the room, even in the presence of intricate geometries.

\begin{table}
  \centering
  \begin{tabular}{lccc}
    \toprule
     & 3D IoU(\%) & CE(\%) & PE(\%) \\
    \midrule
    HorizonNet~\cite{HorizonNet} & 84.61 & 0.65 & \textbf{1.89} \\
    LGT-Net~\cite{LGTNet} & 85.29 & 0.67 & 2.11  \\
    Ours & \textbf{86.19} & \textbf{0.63} & 1.90 \\
    \bottomrule
  \end{tabular}
  \caption{Quantitative results evaluated on Pano\_S2D3D dataset~\cite{PanoContext, S2D3D}.}
  \label{table:cuboid_results}
\end{table}

\begin{table}
  \centering
  \begin{tabular}{lcc}
    \toprule
     & 2D IoU(\%) & 3D IoU(\%) \\
    \midrule
    HorizonNet~\cite{HorizonNet} & 82.51 & 80.04 \\
    LED$^{2}$-Net~\cite{Led2Net} & 82.87 & 80.62 \\
    LGT-Net~\cite{LGTNet} & 83.69 & 81.21 \\
    Ours & \textbf{83.93} & \textbf{81.62} \\
    \bottomrule
  \end{tabular}
  \caption{Quantitative results evaluated on Matterport3D dataset\cite{Mt3D}.}
  \label{table:overall_performance}
\end{table}

\begin{figure}[t]
  \centering
   \includegraphics[width=1.0\linewidth]{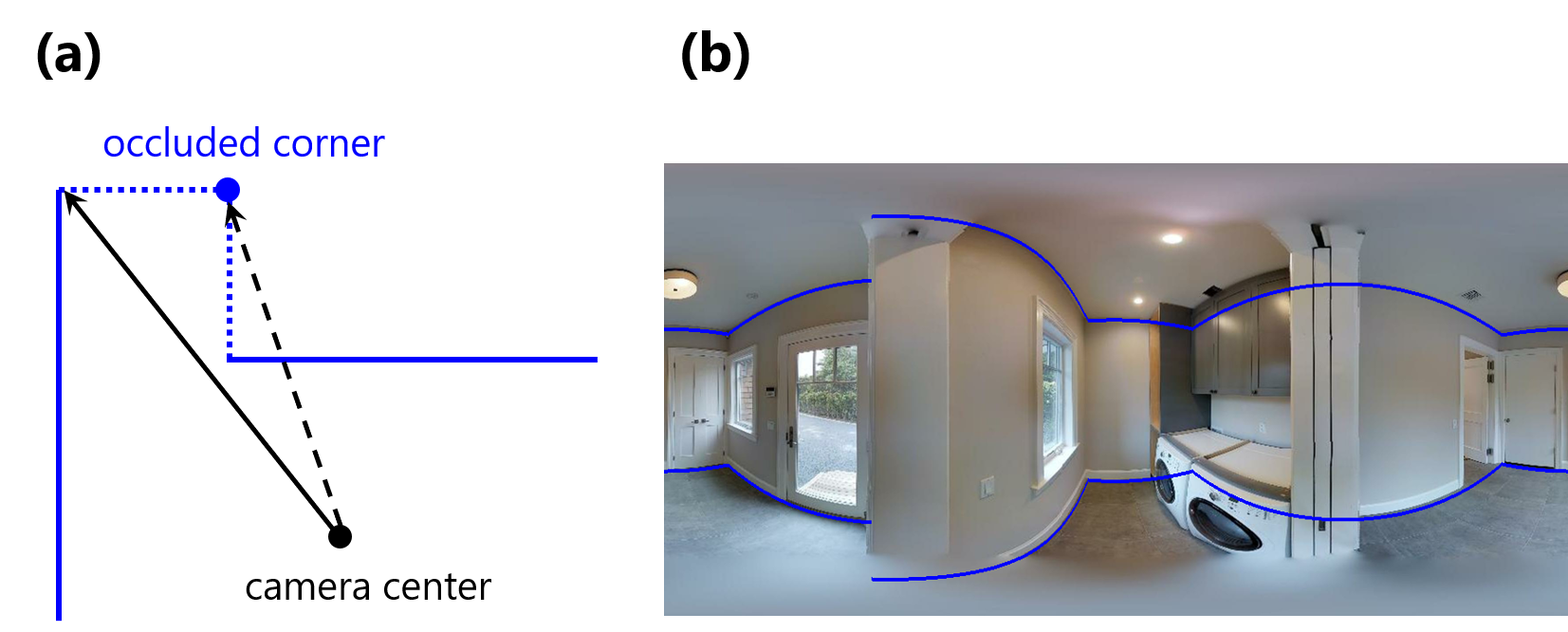}

   \caption{(a) Illustration of occlusion occurring in room. (b) Image when occlusion occurs. Wall boundaries colored in blue line are discontinuous where occlusion occurs in image.}
   \label{fig:occlusion}
\end{figure}

\begin{figure*}
\begin{center}
\includegraphics[width=1.0\linewidth]{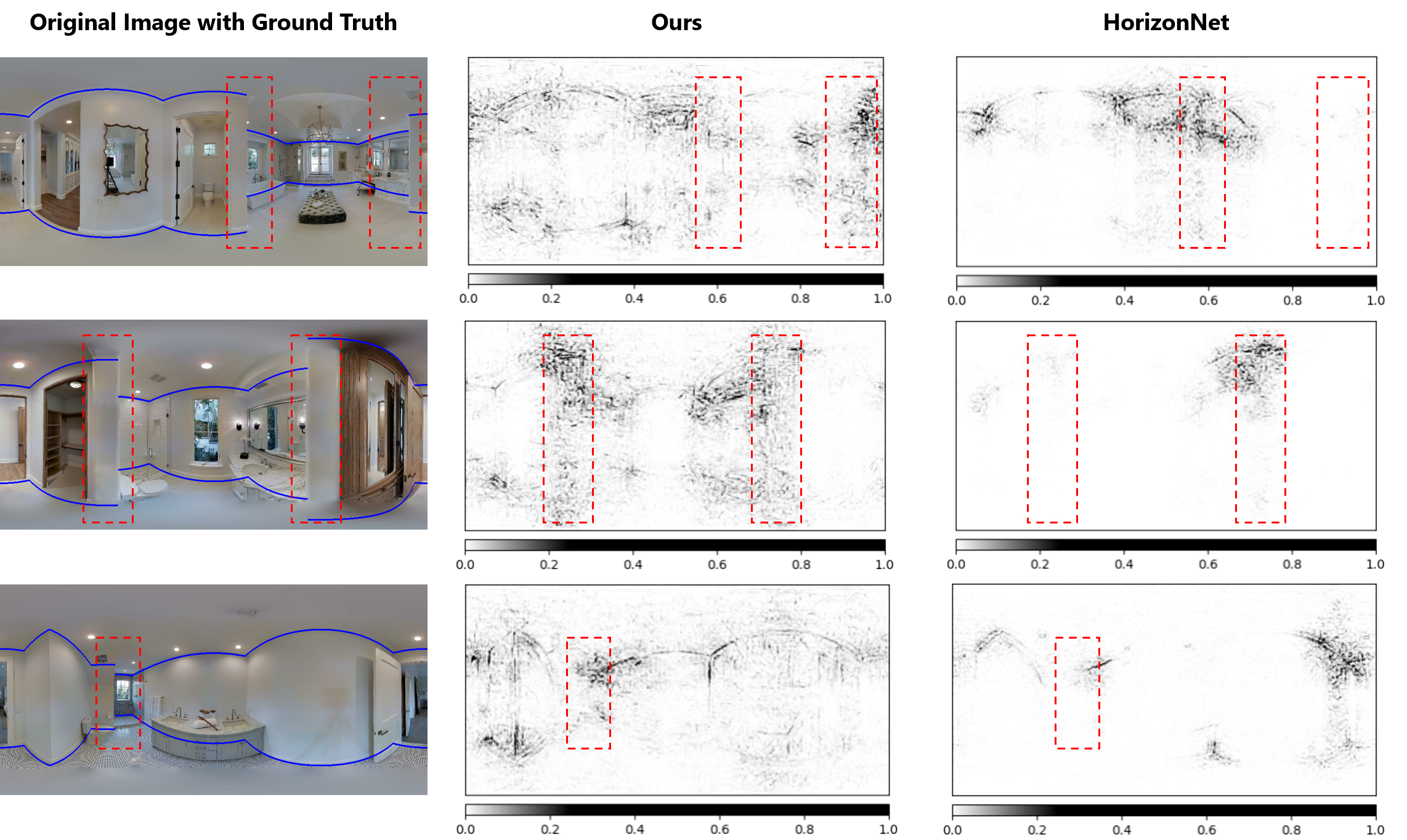}
\end{center}
   \caption{
   Visualization of Integrated Gradients of each model. Original images are aligned in left column with ground truth of wall boundaries colored in blue. Color bar at bottom of images in two columns on right side shows attribution magnitude; darker color shows higher attribution magnitude. Areas inside red dashed rectangles indicate where occlusion occurs.
   }
\label{fig:ig}
\end{figure*}

\subsection{Results on Occlusion Dataset}
To verify the robustness of our model to occlusion, we evaluated the models on 132 images with occlusion from the test split of the Matteport3D dataset~\cite{Mt3D}. In this experiment, occlusion was defined as a situation where one or more corners of the ground truth are not visible, as illustrated in \cref{fig:occlusion}. The Pano\_S2D3D dataset~\cite{PanoContext, S2D3D} was not included in this experiment because of its cuboid nature, which does not cause occlusion. Quantitative results on the occlusion dataset are presented in \cref{table:occlusion_dataset}.
We reported the best performance of our model, a 2.37\% improvement in both 2D and 3D IoU compared with LGT-Net~\cite{LGTNet}. In comparison to \cref{table:overall_performance}, the difference in 2D and 3D IoU between our model and other models was greater on the occlusion dataset. This result suggests that our model is more resistant to occlusion.

To visualize the difference in the results, we used Integrated Gradients: a method for obtaining the contribution of input elements to the output~\cite{integrated_gradients}. As shown in \cref{eq:ig}, we integrate gradients at all points along a linear path from baseline $x' \in \mathbb{R}^{N}$ to input $x \in \mathbb{R}^{N}$. We implemented the visualization of Integrated Gradients using Captum~\cite{captum}.
\begin{multline}
  Integrated\_Gradients_{i}(x) \\
  = (x_{i} - x'_{i})\int_{0}^{1}\frac{\partial F(x'+\alpha(x-x'))}{\partial x_{i}}d\alpha
  \label{eq:ig}
\end{multline}

For visibility, we compared our model with HorizonNet~\cite{HorizonNet}, which estimates corners as ours do. The results of Integrated Gradients of each model are depicted in \cref{fig:ig}. The black areas had a high magnitude of Integrated Gradients and were more attended by the model. The area enclosed by the red dashed line in our model is darker than in HorizonNet~\cite{HorizonNet}, indicating that our model focused more on occluded regions. This suggests that the difference in model attention causes the proposed model to be more resistant to occlusion. Furthermore, our model showed diffused black areas, which indicates that it has a more global attention. This may contribute to increasing in the model's accuracy on overall data.

\begin{table}
  \centering
  \begin{tabular}{lcc}
    \toprule
     & 2D IoU(\%) & 3D IoU(\%) \\
    \midrule
    HorizonNet~\cite{HorizonNet} & 75.92 & 73.30 \\
    LED$^{2}$-Net~\cite{Led2Net} & 75.65 & 73.64 \\
    LGT-Net~\cite{LGTNet} & 76.92 & 75.03 \\
    Ours & \textbf{78.74} & \textbf{76.81} \\
    \bottomrule
  \end{tabular}
  \caption{Quantitative results evaluated on occluded data in test split of Matterport3D dataset\cite{Mt3D}.}
  \label{table:occlusion_dataset}
\end{table}

\subsection{Ablation Studies}
We performed ablation studies to assess the effectiveness of Cutout (CUT), knowledge distillation (KD), and IoU loss (IoU) on the Matterport3D dataset~\cite{Mt3D}. The results are presented in \cref{table:ab_study}. Model 1 in the index lacks all the mentioned methods, while model 8 in the index incorporates all of them. A comparison of models at indices 2-4 with the model at index 1 demonstrates that each method contributed independently to improving the model accuracy. Cutout shows the smallest effect of the three methods. Notably, the integrating of knowledge distillation and IoU loss (indexed 7) yielded a significant improvement in accuracy. It indicates that the layout information provided by knowledge distillation can be utilized effectively to reduce IoU loss in areas where occlusion occurs. On the other hand, the lower accuracy of models indexed 5 and 6 compared with those indexed 3 and 4 can be attributed to the inability of knowledge distillation and IoU loss alone to handle pseudo-increased occlusion generated by Cutout. Model 8 in the index remarked the highest accuracy of all, which suggests both knowledge distillation and IoU loss can deal with the pseudo-occlusion image. The difference of models indexed 7 and 8 was relatively small, indicating Cutout plays a supplemental role in layout estimation. These results confirm that the combination knowledge distillation and IoU loss is effective, whereas Cutout augmentation serves as an ancillary method for layout estimation.

\begin{table}
  \centering
  \begin{tabular}{cccccc}
    \toprule
    & \multicolumn{3}{c}{Method} & \multicolumn{2}{c}{Metrics}  \\
    \hline
    Index & CUT & KD & IoU & 2D IoU(\%) & 3D IoU(\%) \\
    \midrule
    1 & & & & 82.53 & 80.22 \\
    2 & \checkmark & & & 82.88 & 80.51 \\
    3 & & \checkmark & & 83.39 & 81.13 \\
    4 & & & \checkmark & 83.41 & 81.10  \\
    5 & \checkmark & \checkmark & & 82.73 & 80.42 \\
    6 & \checkmark & & \checkmark & 83.07 & 80.69 \\
    7 & & \checkmark & \checkmark & 83.84 & 81.53 \\
    8 & \checkmark & \checkmark & \checkmark & \textbf{83.93} & \textbf{81.62} \\
    \bottomrule
  \end{tabular}
  \caption{Quantitative results of ablation studies evaluated on Matterport3D dataset\cite{Mt3D}. KD, IoU, CUT represents knowledge distillation, IoU loss, Cutout, respectively.}
  \label{table:ab_study}
\end{table}

\subsection{Input Methods}\label{Input}
\begin{figure}[t]
  \centering
   \includegraphics[width=1.0\linewidth]{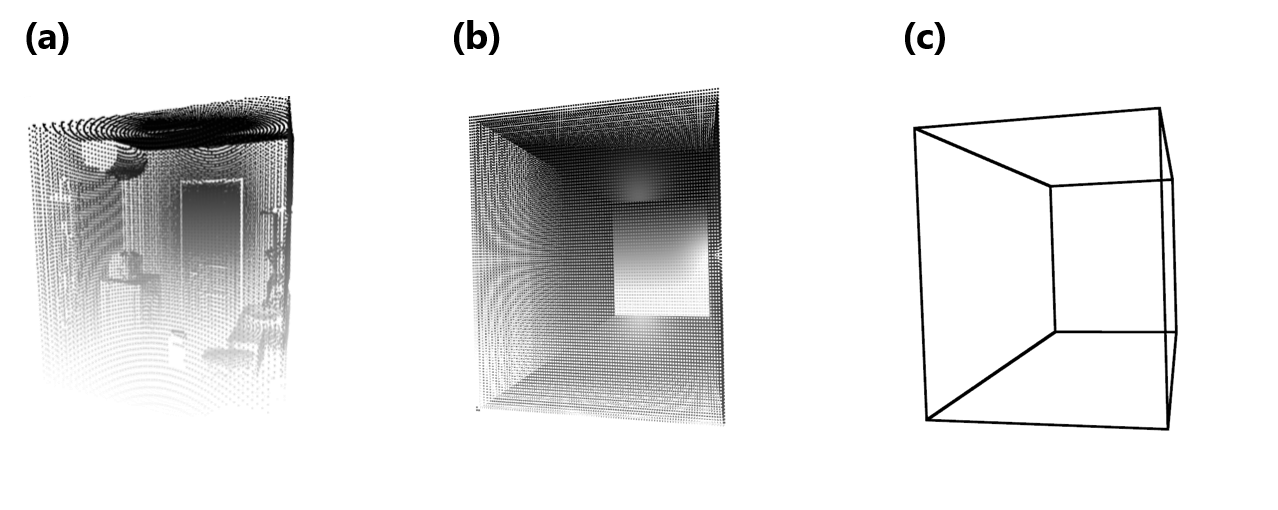}
   \caption{Input point cloud. (a) dense pcd: point cloud including room geometries with furniture generated from depth image. (b) sparse pcd: sparse point cloud including walls generated from annotation on image. (c) layout pcd: point cloud including only wall boundaries generated from annotation on image.}
   \label{fig:pcd}
\end{figure}

We assessed a method for providing 3D input as a point cloud for layout estimation. Typically, 3D information of rooms is acquired using 3D scanners (e.g., LiDAR) or the room geometry data from drawings. We evaluated three types of point clouds for input to layout estimation: dense pcd, sparse pcd, and layout pcd as shown in \cref{fig:pcd}. Dense pcd is a point cloud generated from 3D scans, a depth image in this study, and includes all objects observed in the image. Sparse pcd and layout pcd are generated by converting the annotation data on an image to 3D coordinates and interpolating them. Points are generated on planes of walls, ceilings, and floors for sparse pcd, while only on frames for layout pcd.

For evaluation, we used the Structured3D dataset~\cite{St3D} as it contains both RGB and depth panoramic images. Each model was trained with the data augmentation described in the paper, except Cutout~\cite{Cutout}. It is because we intended to evaluate the effect of the furniture in the dense pcd on layout estimation. The quantitative results in \cref{table:input_methods} demonstrate that the model using layout pcd as input outperformed other models in 2D and 3D IoU. The furniture in the dense pcd occluded some parts of the room geometries, which may have led to reduced accuracy. The results of sparse pcd and layout pcd indicate that walls in the sparse pcd disturb layout estimation. Thus, we verified the layout pcd is the most effective method of feeding point clouds to our model, and adopted layout pcd for the input to our model in the paper.

\begin{table}
  \centering
  \begin{tabular}{lcc}
    \toprule
    Input & 2D IoU(\%) & 3D IoU(\%) \\
    \midrule
    dense pcd & 92.59 & 91.34 \\
    sparse pcd & 92.74 & 91.45 \\
    layout pcd & \textbf{92.83} & \textbf{91.61} \\
    \bottomrule
  \end{tabular}
  \caption{Quantitative results of models with each input evaluated on Structured3D dataset~\cite{St3D}.}
  \label{table:input_methods}
\end{table}

\section{Conclusion}
This paper proposes a novel model that estimates room layouts, taking into consideration the presence of occlusions. To achieve this, we distilled the knowledge from panoramic images and 3D coordinates as inputs, and utilized a 3D IoU loss function. Knowledge distillation allows the model to estimate layouts of rooms even without drawings. Our model outperformed existing models on benchmark datasets, and we demonstrated the robustness of our model to occlusion through evaluation on the dataset of occluded images and visualization of the model's attention. Furthermore, the efficacy of the proposed modules is confirmed through ablation studies.

{\small
\bibliographystyle{ieee_fullname}
\bibliography{main}
}

\end{document}